\pdfoutput=1

\documentclass[11pt]{article}

\usepackage[]{emnlp2021}

\usepackage{times}
\usepackage{latexsym}
\usepackage{amssymb}
\usepackage[T1]{fontenc}

\usepackage[utf8]{inputenc}

\usepackage{microtype}
\usepackage{pifont}
\usepackage{makecell}
\usepackage{lscape}
\usepackage{multirow}
\usepackage{tabularx}
\usepackage{paralist}
\usepackage{algorithm2e}

\newcommand{\centercell}[1]{\multicolumn{1}{|c|}{#1}}
\usepackage{graphicx}
\usepackage{amsmath}
\usepackage{enumitem}
\usepackage{soul, color}
\graphicspath{{figure/}}
\DeclareMathSymbol{*}{\mathbin}{symbols}{"03} 

\title{SciXGen: A Scientific Paper Dataset for Context-Aware Text Generation}
\author{ Hong Chen$^{1,3}$,  Hiroya Takamura$^{2,3}$,  Hideki Nakayama$^{1,3}$\\ The University of Tokyo$^1$, Tokyo Institute of Technology$^2$\\ National Institute of Advanced Industrial Science and Technology, Japan$^3$  \\  \texttt{\{chen, nakayama\}@nlab.ci.i.u-tokyo.ac.jp}\\ \texttt{takamura.hiroya@aist.go.jp}
}

\begin{document}
\maketitle
\begin{abstract}

Generating texts in scientific papers requires not only capturing the content contained within the given input but also frequently acquiring the external information called \textit{context}.
We push forward the scientific text generation by proposing a new task, namely \textbf{context-aware text generation} in the scientific domain, aiming at exploiting the contributions of context in generated texts.
To this end, we present a novel challenging large-scale \textbf{Sci}entific Paper Dataset for Conte\textbf{X}t-Aware Text \textbf{Gen}eration (SciXGen), consisting of well-annotated 205,304 papers with full references to widely-used objects (e.g., tables, figures, algorithms) in a paper.
We comprehensively benchmark, using state-of-the-arts, the efficacy of our newly constructed SciXGen dataset in generating description and paragraph.
Our dataset and benchmarks will be made publicly available to hopefully facilitate the scientific text generation research.

\end{abstract} 
\section{Introduction}

\begin{table}[tb]
\footnotesize
\begin{tabular}{|p{7cm}|}
\hline
\centercell{\textbf{Body text (Context)}}
\\\hline
$\ldots$ {\color{blue}languages}: Telugu (te) and Turkish (tr)$\ldots$ Turkish (tr) vocabulary has been censored to {\color{blue}contain no overlap} with the Telugu$\ldots$ we evaluate these models using a {\color{blue}recall@k metric} defined as $\ldots$ 
\\\hline
\centercell{\textbf{Table}}
\\\hline
\centercell{    \begin{tabular}{|l|c|c|c|}

     Result    &  \textbf{te+en}  &  + \textbf{tr}  &  \% Change \\\hline
     Recall@1  &  17.0  &  17.6  &  +3.5\% \\
     Recall@10  &  23.9  &  25.0  &  +4.6\% \\
     Recall@20  &  26.3  &  27.7  &  +5.3\% \\
    \end{tabular}}
\\\hline
\centercell{\textbf{Generated description   w/o context (table only)}} 
\\\hline
Table shows when {\color{red}{te+en} is replaced with {tr}}, the effect of different change is very small, although {\color{red}the performance of {tr} method gets really strong}.
\\\hline
\centercell{\textbf{Generated description   w/ context (body text + table)}}
\\\hline
Table summarizes the {\color{blue}recall@1 measures} and the percentage of the incremental improvement {\color{blue}across languages} for both tasks. The average incremental improvement across languages is about 4\% in these cases, {\color{blue}despite there being no overlap between in Telugu and Turkish}.
\\\hline
\end{tabular}
\caption{An example in table description generation (table-to-text) task. Highlighted texts in {\color{red} red} denote the factual incorrectness ({\color{red}hallucination}), and texts in {\color{blue} blue} indicate the fact that can be referred from the context. We can see that tables in the scientific domain contain terms and abbreviations that are mentioned in its body text (i.e. tr and te). With the help of the context, the generated description becomes more plausible.} 
\label{tab:task}
\end{table}

Text generation in the scientific domain has been increasingly received attention recently due to its wide range of applications such as summarization~\cite{lu-etal-2020-multi-xscience}, paragraph generation~\cite{wang2019paperrobot} and table description generation~\cite{moosavi2021learning}. 
Though recent works have brought breakthroughs~\cite{lu-etal-2020-multi-xscience,wang2019paperrobot,moosavi2021learning}, how to faithfully generate texts/paragraphs remains challenging.
As a case study shown in Table~\ref{tab:task}, generating plausible table descriptions always requires not only tabular data itself as the input, but also numerous references to the external information (e.g., body text) as the \textit{context}.
To this end, we promote a new task of \textbf{context-aware text generation} (i.e., generating text given a context), a new branch of text generation research in the scientific domain.
This task can be straightforwardly extended to several specific requirements where we, in this paper, investigate context-aware description generation (i.e., generating description for paper objects such as tables and figures, given the body text as context), and context-aware paragraph generation (i.e., generating a paragraph given cited papers as context).
Therefore, context-aware text generation yields helpful tools to generate scientific papers automatically, yet it has not been well explored in literature.

For conducting experiments on context-aware text generation, a well-developed dataset with complete contextual information is required.
However, existing corpora~\cite{radev2013acl, clement2019arxiv,lo2020s2orc,saier2020unarxive} are not applicable in our task.
~\citet{radev2013acl} and \citet{clement2019arxiv} directly extract data from PDF, failing in capturing the paper structure and other objects (e.g., citations).
Their datasets thus only provide raw texts without any contextual information.
Recently, S2ORC(L{\small A}T{\small E}X)~\cite{lo2020s2orc} and unarXive~\cite{saier2020unarxive} draw out the data from the widely-used format in a scientific paper (i.e., L{\small A}T{\small E}X) to better preserve the paper structure.
S2ORC(L{\small A}T{\small E}X)~\cite{lo2020s2orc} steps further in enriching the data by introducing the references to tables, figures and equations.
Though S2ORC(L{\small A}T{\small E}X)~\cite{lo2020s2orc} contains contextual information to some extent, it still has several limitations:
1) Their tables and figures are not in a machine-readable format.
2) Some objects, such as algorithms and theorems, are not retained.
3) A considerable amount of tables and figures lose content due to their low-quality LaTeX parser.

Inspired by the above observations, we introduce a novel large-scale scientific paper dataset (SciXGen) designed explicitly for context-aware text generation.
Our dataset consists of 205,304 papers with references to all types of objects in a paper, yielding fully complete contextual information.
We believe our dataset can be served as a testbed for research and evaluation on the task of text generation in the scientific domain.
We also provide a benchmark to illustrate the importance of contextual information in the text generation problem.
More precisely, we evaluate several state-of-the-arts in context-aware description and paragraph generation tasks under various scenarios.

Our contributions can be summarized as follows:

\begin{itemize}
[leftmargin=0cm,itemindent=.5cm,labelwidth=\itemindent,labelsep=0cm,align=left]
    \item We address a new task of context-aware text generation in the scientific domain.
    To the best of our knowledge, this paper is the first fully exploring the contributions of context in scientific text generation. 
    In addition, we define two primary tasks in this problem: context-aware description generation and context-aware paragraph generation.
    \item We introduce a novel large-scale challenging dataset (SciXGen) to promote context-aware text generation research. 
    Samples in our dataset can be found in  \url{paperparser.com/display}\footnote{This website is anonymous for double-blind reviewing at the time of submission.}.
    \item We provide benchmarks for context-aware description generation and context-aware paragraph generation tasks. In particular, we slightly modify state-of-the-art methods to adapt the requirements of these tasks and extensively evaluate the models under various scenarios.
\end{itemize}

\section{Related Work}
\label{sec:relate}
\subsection{Datasets of scientific papers}
Existing datasets of scientific papers can be classified roughly into two groups: 
\textit{corpus-level datasets} and \textit{task-specific datasets}. 
The former group often uses a PDF parser to draw out the raw texts from a paper. 
ACL Anthology~\cite{radev2013acl}, Arxiv CS~\cite{clement2019arxiv} and PubMed\footnote{https://www.ncbi.nlm.nih.gov/pmc/tools/openftlist/} consist of 25K, 90K and 2.6M papers, contributing to computational linguistics, computer science and biomedical, respectively.
However, these datasets do not contain the citations, equations and paper structures due to the limitation of the PDF parser.

Recently, unarXive~\cite{saier2020unarxive} and S2ORC~\cite{lo2020s2orc} parses 1.5M papers from their source (i.e., L{\small A}T{\small E}X), providing the possibility to deal with different types of objects (e.g., tables, figures and more).
On the other hand, task-specific datasets are tailored for specific tasks, such as paraphrase generation~\cite{dong-etal-2021-parasci}, summarization~\cite{lu-etal-2020-multi-xscience} and table-to-text~\cite{moosavi2021learning}. 
Most of them are built upon the corpus-level datasets and add task-specific features for different tasks.

In this paper, we propose SciXGen, a corpus-level dataset, which parses the body text more precisely to retain more information from the papers. 
Thus, task-specific datasets can be easily obtained for different tasks in context-aware text generation.

\subsection{Text generation in scientific domain}
Text generation in the scientific domain has achieved progress in several ways.
~\citet{wang2019paperrobot} generates the paper abstract from the input title along with predicted entities in the related papers and further generates the paragraphs for the conclusion and future work.
~\citet{demir2019neural} generates the L{\small A}T{\small E}X source code with a sequence-to-sequence model in a straightforward manner. 
~\citet{lu-etal-2020-multi-xscience} and \citet{an2021enhancing} summarize the abstracts from cited papers to generate a paragraph for related works.
~\citet{moosavi2021learning} learns from the tables in the paper to generate the textual description. 
However, none of aforementioned works make full use of the contextual information within the papers. 
In this work, we propose context-aware text generation tasks that allow the model to access complete contextual information, which is more similar to reading and writing papers.

\section{The SciXGen Dataset}

\subsection{Dataset construction}
\label{sec:construction}

\begin{table*}[tb]
\footnotesize
\begin{tabular}{lcccccc}
\hline
Corpus       & Papers & Source & \makecell{References\\to objects} & \makecell{Postprocess\\to objects} & \makecell{Linked}     & Scope           \\\hline

AAN~\cite{radev2013acl}                & 25K    & PDF    & none                     & -           & ACL Anthology        & comp ling       \\
arXiv CS~\cite{clement2019arxiv}           & 90K    & PDF    & none                     & -           & arXiv                & cs              \\
CiteSeerX~\cite{huang2015neural}          & 1.0M   & PDF    & none                     & -           & CiteSeerX            & multi           \\
PubMed Central(OA) & 2.3M   & PDF    & partial                  & no          & PubMed               & bio           \\
S2ORC(PDF)~\cite{lo2020s2orc}         & 8.1M   & PDF    & partial                  & no          & S2ORC(full)          & multi  \\
unarXive~\cite{saier2020unarxive}     & 1.5M   & LATEX  & none                     & -           & MAG                  & physics,math,cs \\
S2ORC(L{\small A}T{\small E}X)~\cite{lo2020s2orc}  & 1.5M   & LATEX  & partial                  & no          & S2ORC(full)          & physics,math,cs \\
SciXGen        & 205K   & LATEX  & complete                 & yes         & S2ORC(full) & cs    \\
\hline
\end{tabular}
\caption{Overview of main datasets in scientific domain. 
SciXGen is the first dataset which contains complete references to all objects and all objects are post-processed into machine-readable text.}
\label{tab:dataset}
\vspace*{-0.5\baselineskip}
\end{table*}

This section details our construction process for SciXGen.
As with the previous work (S2ORC(L{\small A}T{\small E}X) and unarXive~\cite{saier2020unarxive}), we construct our SciXGen using the source data from arXiv Bulk Data~\cite{clement2019arxiv}.
We employ 225,495 papers in computer science between 2012.1 and 2020.11 from arXiv, where each paper is in the LaTeX format.
We, in what follows, summarize the main procedure.
It is worth noting that we finally obtained 205,304 high-quality papers out of 225,495 original ones.

\paragraph{Processing LATEX source.}
Since the LaTeX format is not machine-readable, we thus follow S2ORC(L{\small A}T{\small E}X) to first parse the latex format to a machine-friendly one, namely XML.
Unlike S2ORC(L{\small A}T{\small E}X), we employ an up-to-date tool (i.e., LaTeXML\footnote{https://dlmf.nist.gov/LaTeXML/}) which is capable of better recognizing uncommon symbols than that used in S2ORC(L{\small A}T{\small E}X) (i.e., Tralics\footnote{http://www-sop.inria.fr/marelle/tralics/}).

However, we observe that LaTeXML would not work correctly if a paper contains symbols that are not pre-defined in its tools.
To parse the papers more precisely, we introduce an auxiliary latex parser as compensation to LaTeXML.
We remark that the auxiliary LaTeX parser is used when LaTeXML encounters a systematic error or loses objects inside.
Our main parser parses the file in an intermediate format (i.e. XML), while our auxiliary LaTeX parser can directly parse the source file (i.e. LaTeX). Both parsers obtain the body text and extract the objects from each file. 
In addition, we group all the objects into seven classes (see Table~\ref{tab:object} and more details in A.1.1 and A.1.2). 
Our system successfully parses (almost) all objects in the paper, maintaining more valuable details than previous works, as seen later in Section~\ref{sec:SciXGen}.

\paragraph{Linking bibliographies to papers.}
In this step, we link the bibliography entries to the papers with full text. 
This step requires to first extract the information from the bibliography entries (e.g., authors, titles and more), and then link the extracted information to the cited papers with full text.

~\citet{saier2020unarxive} first collected 500 human-annotated bibliography entries from the Cora dataset\footnote{https://relational.fit.cvut.cz/dataset/CORA} as training data.
They then trained an LSTM-based Neural ParsCit~\cite{prasad2018neural} model to recognize and locate the entities such as titles and authors inside the bibliography entries.
However, their data are heavily biased to the old papers, making it difficult of applying their method to recent papers covering a wide range of topics.

Based on the above observation, we incrementally improve the model in ~\citet{saier2020unarxive} by collecting more training data and re-training the model.
To be more specific, we first manually annotate additional 1,500 samples, randomly selected in our dataset.
As a result, we have 2,000 training samples in total.
Next, we replace LSTM in~\cite{saier2020unarxive} by BERT~\cite{devlin2019bert} to better identify named entities from bibliography entries.
As a consequence, we achieve an average accuracy score of 99\% over all the entities.


Next, we resolve the citation links in our data between the papers with full text in S2ORC(full) by matching the author and title information extracted from the bibliography entries to the metadata in S2ORC(full). We use S2ORC(full) as an external database because it is the largest corpus of papers, and contains both full-text data extracted from PDF and LaTeX files.
As a result, S2ORC(full) provides many links between the citations to the full-text papers, thus, providing more fruitful contextual information across the cited papers.

\paragraph{Postprocessing and adding more features.}
\label{para:post}
This step helps to improve the quality of our dataset.
To this end, we transform the objects including tables and figures into a machine-readable format, and highlight the equation.
For tables, we convert the tabular text from a heavily structured XML parsed from LaTeXML into a linear string with special tokens to separate the rows and columns.
For figures, we consistently transform all the figures into PNG format.
For equations, we continue to use the LaTeX format as it is already a machine-friendly format.
Nevertheless, we make our effort to replace the user-defined commands in the equations to minimize the negative consequences of massive symbols.
Moreover, we use \$*\$ to cover the inline equations and special tokens $ \langle equation \rangle$ * $ \langle /equation \rangle$ for the regular ones.
Additionally, we mark out the emphasized words in the content (i.e., bold and italic font) to distinguish them from ordinary words.
We step further in post-processing by filtering out the papers that either lack section information or contain an excessive (>12,000) or insufficient number of (<1,000) words, and finally obtain 205,304 paper data in total.

Besides, we add two additional features for the future research.
1) We first use SPECTOR~\cite{cohan2020specter} on each paper to obtain document-level representations and then identify 300 papers that share similar representations across the dataset.
2) We merge our data with paperwithcode dataset\footnote{paperswithcode.com} to provide links to the original code.

\subsection{Dataset specifications and statistics}
\label{sec:SciXGen}
Our SciXGen contains a total of 205,304 papers, each of which is with an average of 5,296 words. 
Besides, we obtain 484,609 tables, 341,564 figures, 134,253 algorithms and 764,724 theorems.
In the body text, 98.76\% of citations can find references to the bibliography entries, and 41.62\% of them can link to the papers with full texts.
Table~\ref{tab:dataset} summarizes the statistics for some primary datasets in this research community. Despite the relatively small capacity of our dataset, we obtain references to all types of objects that well post-processed for the text generation tasks.
We show the types of objects included in each dataset in Table~\ref{tab:object}. Note that we categorize all types of human-defined objects into seven classes. For example, the object type ``proof'' and ``lemma'' are categorized into the theorem as they share similar content (i.e., words and equations).
\begin{table}[tb]
\centering
\footnotesize
    \begin{tabular}{lccc}
    \hline
    Objects(\%)           &  SciXGen   & \makecell{S2ORC\\(L{\small A}T{\small E}X)} & \makecell{unarXive}\\
    \hline
    Table               & 100.0(+1.9) &  50.1 & -\\
    Figure              & \ \ 93.7(+4.2)  &  59.7 & -\\
    Equation            & 100.0(+0.0) &  99.9 & -\\
    Algorithm           & 100.0(+0.0) &  -    & -\\
    Theorem             & 100.0(+0.0) &  -    & -\\
    Verbatim            & 100.0(+0.0) &  -    & -\\
    Text                & 100.0(+0.0) &  -    & -\\
    \hline
    \end{tabular}
    \caption{The percentage of objects that contain contents. The numbers in the brackets show the improvement of using the auxiliary parser. We can see that S2ORC(L{\small A}T{\small E}X) loses almost half contents in the objects, while SciXGen preserves almost all of them.}
    \label{tab:object}
    \vspace*{-0.3\baselineskip}
\end{table}

Table~\ref{tab:object} also shows the percentage of objects that contain content in the following formats: tabular data for tables, image paths for figures, and text for all other objects. 
For a fair comparison, we compare the datasets using parsed results for same papers.
We see that S2ORC(L{\small A}T{\small E}X) loses nearly half of its content in the form of image paths and tabular data, while SciXGen retains most of them. 
The auxiliary parser further helps us retain 1.9\% tabular data and 4.2\% image path to the figures.
We are currently producing data only in the computer science field, while our LaTeX parser can be applied to any LaTeX sources regardless of the field. We plan to publish the remaining papers later.

\section{Context-Aware Text Generation}
We conduct the experiments on two primary tasks in context-aware text generation: \textit{context-aware description generation} (i.e., generating description for paper objects such as tables, figures, algorithms and theorems given the body text as context), and the \textit{context-aware paragraph generation} (i.e., generating a paragraph given cited papers as context).

\label{sec:task}

\subsection{Context-aware description generation}

Let $x$ denote the content of an object to be described where $x$ can be an image for the figure, a tabular text for the table, or a text for other objects such as algorithm and theorem.
We define the target text as $\tilde{t}$ 
and the context supporting the object as $C$.
Formally, our model receives a tuple of $x$, a token for separation, and $C$ as its input and outputs the target description $\tilde{t}$.


We heuristically estimate $\tilde{t}$ and $C$ as follows.
We use the passage that describes the object as the target $\tilde{t}=c_{i,j}$, where $c_{i,j}$ denotes the passage that begins with the $i$-th sentence and ends with the $j$-th sentence within the body text.
Context then can be $C=c_{0,i-1}$, which is the entire previous texts to the target description.
Our heuristic strategy is based on our two empirical observations.
First, the object description always starts with a sentence that first refers to the object; it ends when reaching the last sentence in the paragraph or encountering a sentence that refers to another object. 
Second, the essential information to describe the objects should be located in the previous text to the target description.
Note that we do not use table/figure captions text as the target since most of them do not provide any in-depth explanation inside the data.

We employ various ways to concatenate the object content $x$ and the context $C$.
To be specific, all objects can be concatenated to $C$ (see Section~\ref{para:post} for postprocessing) except for figures, because they do not have textual information.
Therefore, we use ViT~\cite{dosovitskiy2020image} to obtain the features from the image before concatenation.



\subsection{Context-aware paragraph generation}
One of the primary objectives of scientific paper generation is to assist researchers with paper writing.
To enable the model to generate plausible paragraphs, we introduce context-aware paragraph generation, which is a task that aims to generate paragraphs for the ``Introduction'' section.
Unlike previous works, which generates paragraph using limited information (e.g., abstract)~\cite{wang2019paperrobot, demir2019neural, lu-etal-2020-multi-xscience}, we provide the model with substantial contextual information $C$, which are the body texts in the cited papers. 
For simplicity, we only use the cited papers involved in the ``Introduction'' section and ignore the objects in them.
Thus, the input can be defined as the tuple of the abstract $a$ and the context $C$, while the target $\tilde{t}$ is the  ``Introduction'' paragraph.

\subsection{Dataset split}
We conduct the experiments using the data derived from SciXGen. Table~\ref{tab:subtask} shows statistics of each task.
In the context-aware description generation task, we obtain over 100K data for all the objects except the algorithm from SciXGen (\#num in Table~\ref{tab:subtask}).
Among them, the number of descriptions for the table and the figure are competitive with those in existing datasets in other domains~\cite{parikh2020totto, chen2015microsoft}. 
For figures, we only use the chart and bar images, as an excessive variety of image types would degrade the performance of text generation.
For tables, we exclude those not having any equal number of columns in each row, as extracting alignment information from such tables is quite challenging. For algorithms and theorems, we only retain data with a token count between 200 and 500, as the token count rapidly increases when a theorem or algorithm involves an excessive number of math equations. We also filter out the data less than 30 words in the target sentences, resulting in an average target $\tilde{t}$ length of 71 (\#avg\_out\_len in Table~\ref{tab:subtask}).
Apart from the object, each sample contains approximately 200 sentences (\#cand in Table~\ref{tab:subtask}) from the context $C$ that are used to support the generation.
Due to the high computational costs associated with such large-scale data, we use a random subset of 30,000/5,000/5,000 samples to train/validate/test our model.

In context-aware paragraph generation,
to perform the generation in a fixed domain, we select 39,523 papers in computer vision. We use 30,000 of them for training, 5,000 for testing, and the rest 4,523 for validation. 
Among them, 61.1\% of the cited papers in the ``Introduction'' section can find full-text data, which we believe is adequate for analyzing the quality of using contextual information in paragraph generation. 
As a result, each data contains a target paragraph with an average of 698.61 words and over 200 passages from the cited papers that support the generation.

\begin{table}[tb]
\footnotesize
\centering
\begin{tabular}{lccc}
\hline
Input      & \#num & \#avg\_out\_len & \#cand   \\\hline
Table     & 136K & 74.05             & 199.93  \\
Figure (chart/bar)    & 155K & 76.60             & 179.35  \\
Algorithm & 56K & 67.94             & 227.65  \\
Theorem   & 175K  & 65.00             & 192.44  \\ 
\multicolumn{4}{c}{\dotfill} \\
Abstract & 205K & 698.61            & 221.11*\\
\hline
\end{tabular}
\caption{Split dataset statistics for the context-aware description generation task (first four rows) and the context-aware paragraph generation task (last row). \#num, \#avg\_out\_len and \#cand denotes the number of samples, the average length of the target passages and the number of sentences or passages(*) that support the task as context, respectively.}
\label{tab:subtask}
\vspace*{-1\baselineskip}
\end{table}

\section{Experiment}
\label{sec:exp}
\subsection{Model architectures}
\paragraph{Ordering-sensitive Fusion-in-Decoder (OFiD).}
Fusion-in-Decoder (FiD)~\cite{izacard2020leveraging} is used as the retrieval-augmented model in the context-aware description generation tasks. 
We propose to retrieve appropriate sentences from the context to minimize the input length, thus reduce computational cost.
Normally, the order relation between retrieved sentences is ignored since each sentence in the context is treated independently.
However, the order of the  retrieved sentences is critical for comprehending their semantics when they are retrieved from the context with sentences in order (i.e., previous sentences to the target).
Thus, we propose Ordering-sensitive Fusion-in-Decoder (OFiD) to process sentences independently in the encoder but joint them with their original order to the decoder.
To be more specific, we obtain the retrieved sentences $z \in \text {top-k}(p(\cdot \mid x))$, reorder them according to their initial index in the context and concatenate them with the object $[x;z_{i_1};z_{i_2};\dots;z_{i_k}]$ as input, where ${i_1}\textless{i_2}\textless\ldots\textless{i_k}$. Then the generator attends to the input and generates the description.

One disadvantage of the FiD-based model is that the retriever cannot be updated because of this concatenation process. Therefore, we design a special reward and define the policy gradient as follows:
\setlength{\abovedisplayskip}{10pt}
\setlength{\belowdisplayskip}{10pt}
\begin{gather*}
\nabla_{\theta} J(\theta)=R\left(z_{i}\right) \nabla_{\theta} \log p_{\theta}(z_i \mid x)\\
R(v) = \lVert \sigma(\tilde{t})-\sigma(v) \rVert,
\end{gather*}
where $p_{\theta}(z_i \mid x)$ denotes the probability of selecting a sentence $z_i$ for the object $x$. The reward $R(z_i)$ is the euclidean distance between the sentence embedding $\sigma(\tilde{t})$ and $\sigma(v)$.
As a result, the retriever learns to retrieve the sentences that share similar semantic information with the target $\tilde{t}$.

\paragraph{Retrieval-Augmented generator (RAG-sequence).}
RAG-sequence is used in the context-aware paragraph generation task.
It considers the to-be-retrieved passages (300 words) independently, generating an output sequence for each concatenated inputs (i.e., abstract and retrieved passages) separately and marginalizing over the output generations.
The context in paragraph generation is the full text over different cited papers, in which no order relation retains.
Additionally, the retriever can be automatically updated through the back-propagation of cross-entropy loss from the generator.
As above, we believe that RAG-sequence is an appropriate model for the context-aware paragraph generation task.
The model can be formalized as:
\setlength{\abovedisplayskip}{5pt}
\setlength{\belowdisplayskip}{-5pt}
\begin{gather*}
    p_{\text {RAG-sequence}}(y \mid a)\approx\\
    \sum_{z \in \operatorname{top}-k(p(\cdot \mid a))} p_{\eta}(z \mid a)\prod_{i}^{N} p_{\theta}\left(y_{i} \mid a, z, y_{1: i-1}\right),\\
\end{gather*}
where $p_{\eta}(z|a)$ denotes the retrieval mechanism probability of selecting passages $z$ for the abstract $a$ and the generator outputs the token $y_{i}$, given the abstract $a$, the retrieved passages $z$ and the previous generated tokens $y_{1: i-1}$.

\begin{table*}[tb]
\footnotesize
\begin{tabular}{ll|cccc|cccc}
\hline
object                  & Input          & PPL   & BLEU-4 & METEOR & MOVERS & Fluency   & Faith. & Entail.    & Overall    \\\hline
\multirow{6}{*}{Table}     & $x$             & 21.34 & 0.76   & 15.23  & 0.08   & 2.37 & 1.57  & 1.80 & 1.69 \\
                           & $C$(20)          & 18.29 & 1.33   & 17.52   & 0.11   & \underline{2.52} & 1.39  & 1.98 & 1.51 \\
                           & $x$+$C$(20)        & 15.82 & 1.86   & 17.67   & 0.12   & 2.47 & \underline{1.73}  & 2.05 & 1.75 \\
                           & $x$+$C$(10)+OFiD(10)  & \underline{15.12} & \underline{2.03}   & \underline{18.43}   & \underline{0.13}   & 2.50 & \underline{1.73}  & \underline{2.09} & \underline{1.77} \\
                           & $x$+$C$($\infty$)       & \textbf{13.64} & \textbf{2.39}   & \textbf{18.82}   & \textbf{0.14}   & \textbf{2.60} & \textbf{1.86}  & \textbf{2.13} & \textbf{1.82} \\
                           & Gold           & -     & -      & -      & -         & 2.96 & 2.61  & 2.80 & 2.79 \\\hline
\multirow{6}{*}{Figure}    & $x$              & 28.56 & 0.77   & 13.24  & 0.04   & 2.62 & 1.24  & 1.46 & 1.20 \\
                           & $C$(20)          & 18.54 & 1.98   & 18.68   & 0.12   & 2.66 & 1.22  & 1.88 & 1.54 \\
                           & $x$+$C$(20)        & \underline{16.35} & \underline{2.21}   & \underline{19.62}   & \underline{0.13}   & \textbf{2.70} & \underline{1.70}  & 1.96 & \underline{1.62} \\
                           & $x$+$C$(10)+OFiD(10)  & 17.10 & 2.15   & 19.25   & 0.12   & 2.64 & 1.64  & \textbf{2.04} & 1.60 \\
                           & $x$+$C$($\infty$)      & \textbf{14.61} & \textbf{2.36}   & \textbf{19.70}   & \textbf{0.14}   & \underline{2.68} & \textbf{1.72}  & \underline{2.02} & \textbf{1.65} \\
                           & Gold           & -     & -      & -        & -      & 2.82 & 2.66  & 2.64 & 2.54 \\\hline
\multirow{6}{*}{Algorithm} & $x$              & 14.64 & 2.57   & 15.47 & 0.10   & \underline{2.60} & 1.80  & 1.96 & 1.56 \\
                           & $C$(20)          & 12.87 & 2.45   & 18.56 & 0.12   & 2.58  & 1.62   & 1.84 & 1.58 \\
                           & $x$+$C$(20)        & 11.30 & 2.95   & \underline{18.95}   & 0.13   & \textbf{2.66} & \textbf{2.04}  & \textbf{2.26} & \underline{1.80} \\
                           & $x$+$C$(10)+OFiD(10)  & \underline{11.06} & \underline{2.98}   & 18.45   & \underline{0.14}   & 2.54 & \underline{2.02}  & \underline{2.24} & \textbf{1.86} \\
                           & $x$+$C$($\infty$)      & \textbf{10.47} & \textbf{3.10}   & \textbf{18.97}   & \textbf{0.14}   & 2.54 & 2.01  & 2.08 & 1.72 \\
                           & Gold           & -     & -      & -        & -      & 2.74 & 2.42  & 2.66 & 2.26 \\\hline
\multirow{6}{*}{Theorem}   & $x$              & \ \ 9.69  & 2.34   & 17.59  & 0.13   & 2.34 & 1.84  & 1.88 & 1.50  \\
                           & $C$(20)          & \ \ 8.75  & 2.17   & 18.43  & 0.14   & \textbf{2.64} & 1.84  & 1.94 & 1.70  \\
                           & $x$+$C$(20)       & \ \ \underline{7.53}  & \underline{3.19}   & \underline{19.80}  & \underline{0.16}   & 2.56 & \textbf{2.06}  & \underline{2.08} & \underline{1.86} \\
                           & $x$+$C$(10)+OFiD(10) & \ \ 7.61  & 3.18   & 19.45  & 0.15   & 2.60 & 2.02  & \underline{2.08} & 1.82 \\
                           & $x$+$C$($\infty$)     & \ \ \textbf{6.88}  & \textbf{3.85}   & \textbf{21.07}  & \textbf{0.17}   & \underline{2.60} & \underline{1.98}  & \textbf{2.18} & \textbf{2.00} \\
                           & Gold           & -     & -      & -       & -      & 2.70 & 2.36  & 2.22 & 2.22\\\hline
\end{tabular}
\caption{Evaluation results on context-aware description generation. We both report the scores from automatic metrics and human evaluation. In human evaluation, Faith. and Entail. denote Faithfulness and Entailment, respectively. The human evaluation is rated from 1 to 3, representing the low to high quality. We emphasize the best score and underline the second-best score for each task.}
\label{tab:td-result}
\vspace*{-0.5\baselineskip}
\end{table*}

\subsection{Implementation details}
Longformer-Encoder-Decoder (LED)~\cite{beltagy2020longformer}, which can accept at most 16,327 tokens, is used as the baseline model. For a fair comparison, we also utilize LED as the generator in OFiD and RAG. It receives inputs (i.e., $x$/$a$ and $C$) and outputs the targets (i.e., $\tilde{t}$).
For training all the baseline models, we used the AdamW~\cite{loshchilov2018decoupled} optimizer. 
The learning rate was initialized at 4e-5 and got a linear schedule with warm-up at the first 10,000 iterations. 
We finetune the models in all tasks for 10 epochs with the same random seed, record the evaluation of each epoch and report the best results.
We run the experiments using 4 Nvidia A100 GPU with a batch size of 4.

\subsection{Evaluation metrics}
We use automatic metrics BLEU~\cite{papineni2002bleu}, METEOR~\cite{denkowski2014meteor} and a neural-based metric MoverScore~\cite{zhao2019moverscore}. 
As automatic scores remain tricky for correctly evaluating the text quality, we conduct human evaluation.
As this task requires professional knowledge in computer science, we hire five annotators with a degree in computer science (2 Master students, 2 PhD students, and 1 Postdoc).
We test the performance in terms of Fluency, Faithfulness, Entailment and Overall. \textbf{Fluency} evaluates the language modelling. \textbf{Faithfulness} assesses how relevant the generated texts and the given inputs are. \textbf{Entailment} only evaluates in the context-aware description generation task to show the likelihood that the sentences can be put into the location after the last sentence in the context. \textbf{Overall} is a subjective criterion that shows the preference by annotators.
During the evaluation, we show the annotators the contextual information, the objects (abstract for the context-aware paragraph generation) and the generated texts by different baselines. 
We assign 50 samples in total for each task to the annotators.

\subsection{Experiment on description generation}

\begin{figure}[tb]
    \centering
    \includegraphics[scale=0.22]{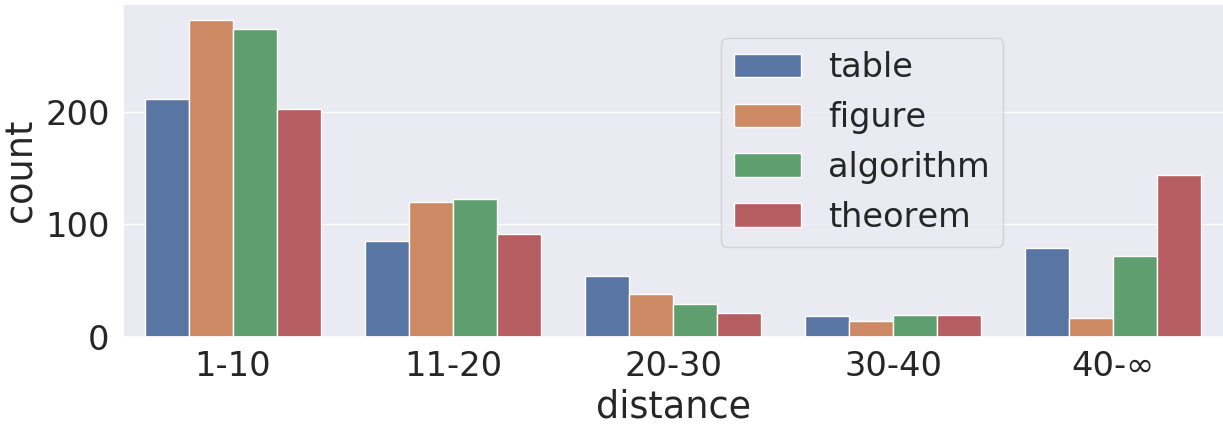}
    \caption{Analysis of possible relevant sentences in the paper. We manually label 10 relevant sentences to the target description from 50 samples for each object. Distance indicates the number of sentences between the selected sentences and the target description.}
    \label{fig:bar}
    \vspace*{-1\baselineskip}
\end{figure}

In this section, we introduce several baselines in context-aware description generation.
First of all, we input the object alone ($x$) into the model to create a baseline without context. 
Then, to determine which sentences in the context are potentially crucial for description generation, we ask the annotators to mark the sentences in the context that would be necessary to infer the target descriptions.
Figure~\ref{fig:bar} shows the results. The distance indicates the number of the sentences between the marked sentences and the target descriptions.
We can see that most of the relevant sentences locate near the target descriptions, which encourages us to use the closest sentences as input context rather than entire sentences in the context to save the computational cost. 
Thus, we propose baselines that use the closest 20 sentences (usually $<$1,000 tokens) to the target descriptions as input ({$C$(20)}) and in conjunction with the object ({$x$+$C$(20)}).
Besides, we keep the 10 closest sentences and use the proposed OFiD to retrieve another 10 sentences from the context ({$x$+$C$(10)+OFiD(10)}).
The final baseline incorporates all sentences in the context ({$x$+$C$($\infty$)}) to show the upper-bound performance of using a pre-trained language model in this task.

\begin{table}[tb]
\centering
\small
\begin{tabular}{l|ccc}
\hline
Precision(\%) & Rand(10) & Dist(11-20) & OFiD(10)   \\\hline
Table     & 5.58   & 12.73    & \textbf{14.83} \\
Figure    & 5.00   & \textbf{12.81}    & \ \ 8.31  \\
Algorithm & 4.39   & 16.75    & \textbf{17.84} \\
Theorem   & 5.20   & \textbf{14.00}    & \ \ 9.31 \\\hline
\end{tabular}
\caption{The sentence retrieval accuracy with different methods. \textbf{Rand(10)} denotes randomly selecting 10 sentences from the context. \textbf{Dist(11-20)} denotes the 11th to 20th sentences before the target description. \textbf{OFiD(10)} denotes the 10 sentences retrieved by OFiD.}
\label{tab:retrieval-acc}
\vspace*{-1\baselineskip}
\end{table}

\begin{table*}[tb]
\footnotesize
\begin{tabular}{l|cccc|cccc}
\hline
Input         & PPL   & BLEU-4 & METEOR  & MOVERS & Fluency & Faithfulness & Overall & Hallucination(\%) \\\hline
$a$      & 10.80 & 5.00   & 32.82  & \textbf{0.18}   & \textbf{2.80}   & 2.50         & 2.23  & 68.58   \\
$a$+RAG(10)      & 10.23 & 5.04   & 32.90  & 0.17   & 2.76    & \textbf{2.62}         & 2.30   &51.28 \\
$a$+RAG(10)$\ast$ & \ \ \textbf{9.67}  & \textbf{5.23}   & \textbf{33.00}  & \textbf{0.18}  & 2.78    & 2.58        & \textbf{2.40}  &\textbf{48.40} \\
\hline
\end{tabular}
\caption{Evaluation results on context-aware paragraph generation.  Same as in the description generation, we both report the scores from automatic metrics and human evaluation.}
\label{tab:para-result}
\vspace*{-1\baselineskip}
\end{table*}

We summarize the results in Table~\ref{tab:td-result}.
We can see that 
1) Baselines that consider only one component ($x$) or $C$(20)) perform worse than those that consider both, indicating that both are critical in text generation in this task.
2) $C$(20) achieving better performance than $x$ reveals that the target descriptions are highly related to the context.
3) In the table and algorithm, retrieval-augmented methods outperform direct use of the closest 20 sentences, while in the figure and theorem, the result shows the contrary.
To analyze the reason, we measure the accuracy of the retrieval by using the same sentences marked in Figure~\ref{fig:bar}.
Table~\ref{tab:retrieval-acc} shows the results. \textbf{Rand(10)} denotes randomly selecting 10 sentences from the context. \textbf{Dist(11-20)} denotes the sentences with indices $i-20$ to $i-11$ (i.e., $c_{i-20, i-11}$), preceeding the target description. \textbf{OFiD(10)} denotes the 10 sentences retrieved by OFiD. The performance gap between $x$+$C$(20) and {$x$+$C$(10)+OFiD(10)} is caused by using different sentences in \textbf{Dist(11-20)} and \textbf{OFiD(10)} as input. From the result, we can deduce that retrieval performance positively affects the quality of text generation, as both the generation and the retrieval have a similar propensity toward performance.
4) Although \textbf{$x$+$C$($\infty$)} achieves the best performance among the baselines, in practice, it requires three times more memory space than other baselines due to its extremely long input size.
As a result, it remains worthwhile to develop a more efficient retrieval approach that improves generation quality while using less memory. Moreover, even if this baseline can attend to all the contextual information, the results are far from perfect, indicating that this task remains challenging.

\subsection{Experiment on paragraph generation}
In this section, we show the results of context-aware paragraph generation. We compare three baselines with different inputs: 1) abstract only ($a$), 2) abstract with additional retrieved 10 abstracts from cited papers ($a$+RAG(10)), 3) abstract with 10 retrieved passages from cited papers ($a$+RAG(10)$\ast$).
Table~\ref{tab:para-result} shows the results. We can see that with the context from cited papers, the generated paragraphs achieve better performance. 
Moreover, compared with the previous work~\cite{lu-etal-2020-multi-xscience, an2021enhancing} that utilizes only the abstract in the cited papers, retrieving passages from full text in cited papers achieves a higher score in automatic metrics and overall scores in human evaluation, thus, indicating the significance of our proposed dataset and tasks.
In addition, we also ask the annotators to measure the hallucination of generated paragraphs. In details, we ask the annotators to check each reference (i.e., given cited papers) whether they are used in the generated paragraph. 
We report hallucination score as the percentage of the papers that have not been mentioned in any place of the generated paragraph (the lower, the better).  
Table~\ref{tab:para-result} indicates that retrieving from context can reduce hallucination from the generator. With more contextual information provided to the model, it can generate more plausible paragraphs with less hallucination.

\subsection{Ablation Studies}
\begin{table*}[t]
\centering
\begin{tabular}{lcccccccc}
\hline
Input      & \#P(M) & \#Mem(GB) & PPL   & BLEU4 & METEOR  & MOVERS \\\hline
Bart-base  & 129    & 6.6       & 14.43 & 2.01   & 17.85   & 0.14   \\
Bart-large & 406    & 16.0      & 14.15 & \textbf{2.43}   & \textbf{20.64}    & \textbf{0.15}   \\
T5-base    & 220    & \ \ 9.6       & \ \ 9.97  & 1.74   & 17.52   & 0.13   \\
T5-large   & 770    & 23.7      & \ \ \textbf{9.11}  & 2.16   & 18.22    & 0.13   \\
LED-base   & 162    & \ *6.8       & 15.82 & 1.86   & 17.67  & 0.12  \\\hline
\end{tabular}
\caption{Automatic results with different pre-trained language models.  We report the parameter numbers and the memory usage for batch size 1 to each model. * means the input token length is limited to 1024.}
\label{tab:lm}
\end{table*}
\subsubsection{Performance of various pre-trained language models}
In the experiment, we use LED-base as the generator for all baselines. 
In this section, we conduct several ablation studies by using different language models. 
We conduct the experiment on context-aware description generation for tables and input {$x$+$C$(20)}).
We test on BART-base~\cite{lewis2019bart}, BART-large, T5-base~\cite{2020t5} and T5-large. 
As shown in Table~\ref{tab:lm}, the language model achieves superior performance using the same model architecture but more parameters(*-base and *-large).
However, the text generation performance varies significantly across different architectures. BART models outperform others in most of the automatic metrics while getting higher perplexity compared with T5.
That might be related to the different corpora that are used during their training.
The inconsistency of perplexity and other automated metrics further points out the drawback of using automatic criteria in these tasks.

\begin{table}[tb]
\footnotesize
\centering
\begin{tabular}{l|llll}\hline
Input          & PPL  & B-4 & M & MS\\\hline
$x$+$C$(10)+OFiD(10)      & \textbf{17.81} &  \textbf{2.03}  & \textbf{18.43}      & \textbf{0.12}       \\
$x$+$C$(10)+FiD(10)       & 18.30 & 1.88  & 17.88   & 0.12 \\\hline     
\end{tabular}
\caption{Results using OFiD and FiD to retrieve sentences. As previously mentioned, OFiD retains the order between sentences, while the original FiD ignores it. PPL, B-4, M and MS denotes perplexity, BLEU-4, METEOR and MoverScore.}
\label{tab:fidrag}
\end{table}

\subsubsection{FiD vs OFiD}
In context-aware description generation, we use OFiD as our retrieval-augmented model. We also compare OFiD with original FiD, which ignores the order information in the context. We take $x$+$C$(20) as our input. As shown in Table~\ref{tab:fidrag}, OFiD outperforms FiD in automatic scores, that proves the order information is critical when retrieving from the context with sentences in order. 



\section{Conclusion}
This paper addresses the novel yet challenging problem of context-aware text generation in the scientific domain.
To promote this task, we present a novel large-scale SciXGen dataset.
We thoroughly investigate the efficacy of our dataset in two primary tasks: context-aware description generation and context-aware paragraph generation.
Despite achieving remarkable results in our experiments, using context in text generation still has room for improvement as above discussions.
We believe that our dataset can serve as a valuable testbed for various tasks in scientific paper research, including summarization with full-text cited papers, and image-text multimodal text generation.





\section{Acknowledgements}
We thank the anonymous reviewers for the useful comments.
This work was supported by JSPS KAKENHI Grant Numbers JP19H04166 and based on results obtained from a project JPNP20006, commissioned by the New Energy and Industrial Technology Development Organization (NEDO).
For experiments, computational resource of AI Bridging Cloud Infrastructure (ABCI) provided by National Institute of Advanced Industrial Science and Technology (AIST) was used.

\bibliography{custom}

\begin{thebibliography}{25}
\expandafter\ifx\csname natexlab\endcsname\relax\def\natexlab#1{#1}\fi

\bibitem[{An et~al.(2021)An, Zhong, Chen, Wang, Qiu, and
  Huang}]{an2021enhancing}
Chenxin An, Ming Zhong, Yiran Chen, Danqing Wang, Xipeng Qiu, and Xuanjing
  Huang. 2021.
\newblock Enhancing scientific papers summarization with citation graph.

\bibitem[{Beltagy et~al.(2020)Beltagy, Peters, and
  Cohan}]{beltagy2020longformer}
Iz~Beltagy, Matthew~E Peters, and Arman Cohan. 2020.
\newblock Longformer: The long-document transformer.
\newblock \emph{arXiv preprint arXiv:2004.05150}.

\bibitem[{Chen et~al.(2015)Chen, Fang, Lin, Vedantam, Gupta, Doll{\'a}r, and
  Zitnick}]{chen2015microsoft}
Xinlei Chen, Hao Fang, Tsung-Yi Lin, Ramakrishna Vedantam, Saurabh Gupta, Piotr
  Doll{\'a}r, and C~Lawrence Zitnick. 2015.
\newblock Microsoft coco captions: Data collection and evaluation server.
\newblock \emph{arXiv preprint arXiv:1504.00325}.

\bibitem[{Clement et~al.(2019)Clement, Bierbaum, O'Keeffe, and
  Alemi}]{clement2019arxiv}
Colin~B. Clement, Matthew Bierbaum, Kevin~P. O'Keeffe, and Alexander~A. Alemi.
  2019.
\newblock \href {http://arxiv.org/abs/1905.00075} {On the use of arxiv as a
  dataset}.

\bibitem[{Cohan et~al.(2020)Cohan, Feldman, Beltagy, Downey, and
  Weld}]{cohan2020specter}
Arman Cohan, Sergey Feldman, Iz~Beltagy, Doug Downey, and Daniel~S Weld. 2020.
\newblock Specter: Document-level representation learning using
  citation-informed transformers.
\newblock In \emph{Proceedings of the 58th Annual Meeting of the Association
  for Computational Linguistics}, pages 2270--2282.

\bibitem[{Demir et~al.(2019)Demir, Mutlu, and {\"O}zdemir}]{demir2019neural}
Samet Demir, Uras Mutlu, and {\"O}zgur {\"O}zdemir. 2019.
\newblock Neural academic paper generation.
\newblock \emph{arXiv preprint arXiv:1912.01982}.

\bibitem[{Denkowski and Lavie(2014)}]{denkowski2014meteor}
Michael Denkowski and Alon Lavie. 2014.
\newblock Meteor universal: Language specific translation evaluation for any
  target language.
\newblock In \emph{Proceedings of the ninth workshop on statistical machine
  translation}, pages 376--380.

\bibitem[{Devlin et~al.(2019)Devlin, Chang, Lee, and
  Toutanova}]{devlin2019bert}
Jacob Devlin, Ming-Wei Chang, Kenton Lee, and Kristina Toutanova. 2019.
\newblock Bert: Pre-training of deep bidirectional transformers for language
  understanding.
\newblock In \emph{Proceedings of the 2019 Conference of the North American
  Chapter of the Association for Computational Linguistics: Human Language
  Technologies, Volume 1 (Long and Short Papers)}, pages 4171--4186.

\bibitem[{Dong et~al.(2021)Dong, Wan, and Cao}]{dong-etal-2021-parasci}
Qingxiu Dong, Xiaojun Wan, and Yue Cao. 2021.
\newblock \href {https://www.aclweb.org/anthology/2021.eacl-main.33}
  {{P}ara{SCI}: A large scientific paraphrase dataset for longer paraphrase
  generation}.
\newblock In \emph{Proceedings of the 16th Conference of the European Chapter
  of the Association for Computational Linguistics: Main Volume}, pages
  424--434, Online. Association for Computational Linguistics.

\bibitem[{Dosovitskiy et~al.(2020)Dosovitskiy, Beyer, Kolesnikov, Weissenborn,
  Zhai, Unterthiner, Dehghani, Minderer, Heigold, Gelly
  et~al.}]{dosovitskiy2020image}
Alexey Dosovitskiy, Lucas Beyer, Alexander Kolesnikov, Dirk Weissenborn,
  Xiaohua Zhai, Thomas Unterthiner, Mostafa Dehghani, Matthias Minderer, Georg
  Heigold, Sylvain Gelly, et~al. 2020.
\newblock An image is worth 16x16 words: Transformers for image recognition at
  scale.
\newblock \emph{arXiv preprint arXiv:2010.11929}.

\bibitem[{Huang et~al.(2015)Huang, Wu, Liang, Mitra, and
  Giles}]{huang2015neural}
Wenyi Huang, Zhaohui Wu, Chen Liang, Prasenjit Mitra, and C~Giles. 2015.
\newblock A neural probabilistic model for context based citation
  recommendation.
\newblock In \emph{Proceedings of the AAAI Conference on Artificial
  Intelligence}, volume~29.

\bibitem[{Izacard and Grave(2020)}]{izacard2020leveraging}
Gautier Izacard and Edouard Grave. 2020.
\newblock Leveraging passage retrieval with generative models for open domain
  question answering.
\newblock \emph{arXiv preprint arXiv:2007.01282}.

\bibitem[{Lewis et~al.(2019)Lewis, Liu, Goyal, Ghazvininejad, Mohamed, Levy,
  Stoyanov, and Zettlemoyer}]{lewis2019bart}
Mike Lewis, Yinhan Liu, Naman Goyal, Marjan Ghazvininejad, Abdelrahman Mohamed,
  Omer Levy, Ves Stoyanov, and Luke Zettlemoyer. 2019.
\newblock Bart: Denoising sequence-to-sequence pre-training for natural
  language generation, translation, and comprehension.
\newblock \emph{arXiv preprint arXiv:1910.13461}.

\bibitem[{Lo et~al.(2020)Lo, Wang, Neumann, Kinney, and Weld}]{lo2020s2orc}
Kyle Lo, Lucy~Lu Wang, Mark Neumann, Rodney Kinney, and Daniel~S Weld. 2020.
\newblock S2orc: The semantic scholar open research corpus.
\newblock In \emph{Proceedings of the 58th Annual Meeting of the Association
  for Computational Linguistics}, pages 4969--4983.

\bibitem[{Loshchilov and Hutter(2018)}]{loshchilov2018decoupled}
Ilya Loshchilov and Frank Hutter. 2018.
\newblock Decoupled weight decay regularization.
\newblock In \emph{International Conference on Learning Representations}.

\bibitem[{Lu et~al.(2020)Lu, Dong, and Charlin}]{lu-etal-2020-multi-xscience}
Yao Lu, Yue Dong, and Laurent Charlin. 2020.
\newblock \href {https://doi.org/10.18653/v1/2020.emnlp-main.648}
  {Multi-{XS}cience: A large-scale dataset for extreme multi-document
  summarization of scientific articles}.
\newblock In \emph{Proceedings of the 2020 Conference on Empirical Methods in
  Natural Language Processing (EMNLP)}, pages 8068--8074, Online. Association
  for Computational Linguistics.

\bibitem[{Moosavi et~al.(2021)Moosavi, R{\"u}ckl{\'e}, Roth, and
  Gurevych}]{moosavi2021learning}
Nafise~Sadat Moosavi, Andreas R{\"u}ckl{\'e}, Dan Roth, and Iryna Gurevych.
  2021.
\newblock Learning to reason for text generation from scientific tables.
\newblock \emph{arXiv preprint arXiv:2104.08296}.

\bibitem[{Papineni et~al.(2002)Papineni, Roukos, Ward, and
  Zhu}]{papineni2002bleu}
Kishore Papineni, Salim Roukos, Todd Ward, and Wei-Jing Zhu. 2002.
\newblock Bleu: a method for automatic evaluation of machine translation.
\newblock In \emph{Proceedings of the 40th annual meeting of the Association
  for Computational Linguistics}, pages 311--318.

\bibitem[{Parikh et~al.(2020)Parikh, Wang, Gehrmann, Faruqui, Dhingra, Yang,
  and Das}]{parikh2020totto}
Ankur Parikh, Xuezhi Wang, Sebastian Gehrmann, Manaal Faruqui, Bhuwan Dhingra,
  Diyi Yang, and Dipanjan Das. 2020.
\newblock Totto: A controlled table-to-text generation dataset.
\newblock In \emph{Proceedings of the 2020 Conference on Empirical Methods in
  Natural Language Processing (EMNLP)}, pages 1173--1186.

\bibitem[{Prasad et~al.(2018)Prasad, Kaur, and Kan}]{prasad2018neural}
Animesh Prasad, Manpreet Kaur, and Min-Yen Kan. 2018.
\newblock Neural parscit: a deep learning-based reference string parser.
\newblock \emph{International Journal on Digital Libraries}, 19(4):323--337.

\bibitem[{Radev et~al.(2013)Radev, Muthukrishnan, Qazvinian, and
  Abu-Jbara}]{radev2013acl}
Dragomir~R Radev, Pradeep Muthukrishnan, Vahed Qazvinian, and Amjad Abu-Jbara.
  2013.
\newblock The acl anthology network corpus.
\newblock \emph{Language Resources and Evaluation}, 47(4):919--944.

\bibitem[{Raffel et~al.(2020)Raffel, Shazeer, Roberts, Lee, Narang, Matena,
  Zhou, Li, and Liu}]{2020t5}
Colin Raffel, Noam Shazeer, Adam Roberts, Katherine Lee, Sharan Narang, Michael
  Matena, Yanqi Zhou, Wei Li, and Peter~J. Liu. 2020.
\newblock \href {http://jmlr.org/papers/v21/20-074.html} {Exploring the limits
  of transfer learning with a unified text-to-text transformer}.
\newblock \emph{Journal of Machine Learning Research}, 21(140):1--67.

\bibitem[{Saier and F{\"a}rber(2020)}]{saier2020unarxive}
Tarek Saier and Michael F{\"a}rber. 2020.
\newblock unarxive: a large scholarly data set with publications’ full-text,
  annotated in-text citations, and links to metadata.
\newblock \emph{Scientometrics}, 125(3):3085--3108.

\bibitem[{Wang et~al.(2019)Wang, Huang, Jiang, Knight, Ji, Bansal, and
  Luan}]{wang2019paperrobot}
Qingyun Wang, Lifu Huang, Zhiying Jiang, Kevin Knight, Heng Ji, Mohit Bansal,
  and Yi~Luan. 2019.
\newblock Paperrobot: Incremental draft generation of scientific ideas.
\newblock In \emph{Proceedings of the 57th Annual Meeting of the Association
  for Computational Linguistics}, pages 1980--1991.

\bibitem[{Zhao et~al.(2019)Zhao, Peyrard, Liu, Gao, Meyer, and
  Eger}]{zhao2019moverscore}
Wei Zhao, Maxime Peyrard, Fei Liu, Yang Gao, Christian~M Meyer, and Steffen
  Eger. 2019.
\newblock Moverscore: Text generation evaluating with contextualized embeddings
  and earth mover distance.
\newblock \emph{arXiv preprint arXiv:1909.02622}.

\end{thebibliography}
\bibliographystyle{acl_natbib}



\end{document}